\pdfoutput=1

\documentclass[11pt]{article}

\usepackage[table]{xcolor}
\usepackage{url}

\usepackage[]{acl}

\usepackage{times}
\usepackage{latexsym}

\usepackage[T1]{fontenc}

\usepackage[utf8]{inputenc}

\usepackage{svg}
\usepackage{color}
\usepackage{tabularx}
\usepackage{tabu}
\usepackage{booktabs}
\usepackage{multirow}
\usepackage{makecell}
\usepackage{graphicx} 
\usepackage{float}
\usepackage{subfigure} 
\usepackage{amssymb}

\usepackage{hyperref}
\usepackage{tablefootnote}
\usepackage{xspace}

\usepackage{float}
\usepackage{amsmath}
\usepackage{bm}
\usepackage{algorithmicx}
\usepackage{algorithm}
\usepackage{algpseudocode}

\usepackage{arydshln}

\usepackage{amssymb}
\usepackage{pifont}
\usepackage{enumitem}

\usepackage{microtype}

\newcommand{\ours}{\textsc{IfIR}\xspace}
\newcommand{\ourmetric}{\textsc{InstFol}\xspace}
\newcommand{\ourmetrick}{\textsc{InstFol@20}\xspace}
\newcommand{\eg}{\hbox{\emph{e.g.,}}\xspace}
\newcommand{\ie}{\hbox{\emph{i.e.,}}\xspace}
\newcommand{\followir}{\textsc{FollowIR}\xspace}
\newcommand{\instructir}{\textsc{InstructIR}\xspace}
\newcommand{\instructor}{\textsc{InstructOR}\xspace}
\newcommand{\bright}{\textsc{BRIGHT}\xspace}

\newcommand{\nexample}{2,426\xspace}
\newcommand{\nmodel}{15\xspace}
\newcommand{\lawfield}{law}

\newcommand{\github}{\raisebox{-1.5pt}{\includegraphics[height=1.05em]{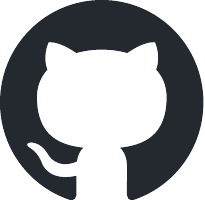}\xspace}}

\title{\ours: A Comprehensive Benchmark for Evaluating \\Instruction-Following in Expert-Domain Information Retrieval}

\author{Tingyu Song$^\clubsuit$ \quad Guo Gan$^\diamondsuit$ \quad Mingsheng Shang$^{\heartsuit \clubsuit}$ \quad Yilun Zhao$^\spadesuit$ \vspace{5pt} \\
    {\small\fontsize{11.2pt}{12pt}\selectfont $^\clubsuit$ School of Advanced Interdisciplinary Sciences, University of Chinese Academy of Sciences} \\
    {\small\fontsize{10pt}{12pt}\selectfont $^\heartsuit$ Chongqing Institute of Green and Intelligent Technology, Chinese Academy of Sciences} \\
    {\small\fontsize{10pt}{12pt}\selectfont $^\diamondsuit$ Zhejiang University} \quad
    {\small\fontsize{10pt}{12pt}\selectfont $^\spadesuit$ Yale University} \vspace{5pt}\\
\github ~\url{https://github.com/SighingSnow/IFIR}
}

\begin{document}
\maketitle
\begin{abstract}
We introduce \ours, the first comprehensive benchmark designed to evaluate instruction-following information retrieval in expert domains. 
\ours includes \nexample high-quality examples and covers eight subsets across four specialized domains: finance, \lawfield, healthcare, and science literature.
Each subset addresses one or more domain-specific retrieval tasks, replicating real-world scenarios where customized instructions are critical.
\ours enables a detailed analysis of instruction-following retrieval capabilities by incorporating instructions at different levels of complexity. 
We also propose a novel LLM-based evaluation method to provide a more precise and reliable assessment of model performance in following instructions.
Through extensive experiments on \nmodel frontier information retrievers, including those based on LLMs, our results reveal that current models face significant challenges in effectively following complex, domain-specific instructions. 
We further provide in-depth analyses to highlight these limitations, offering insights to guide future advancements in retriever development.

\end{abstract}

\begin{figure}[!t]
    \centering
    \includegraphics[width=0.97\linewidth]{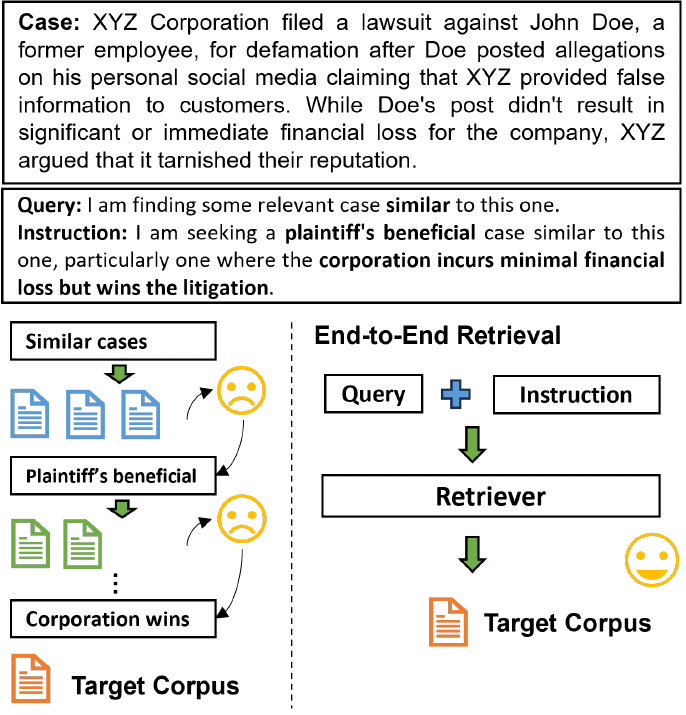}
    \caption{
(Top): An illustration of instruction-following IR scenarios explored in this study. The example simulates a legal case search, where the user provides detailed instructions to retrieve relevant legal cases. Current IR systems struggle to handle such complex queries. (Bottom left): As a result, users have to break down their information needs into simpler, iterative search queries and manually filter the retrieved cases, resulting in a time-consuming and inefficient process.
(Bottom right): This study focuses on evaluating the progress and limitations of current end-to-end retrieval systems in expert-domain instruction-following IR.
    }
    \label{fig:complex-instruction}
\end{figure}
\section{Introduction}
The instruction-following ability has become a cornerstone for LLMs~\cite{ouyang2022training, jiang2023mistral, groeneveld-etal-2024-olmo, dubey2024llama3, yang2024qwen2}, empowering them to interpret and respond to complex user commands and perform a wide range of user-specific tasks. 
Despite its critical importance, the instruction-following capability remains underexplored in the context of information retrieval (IR). 

Current information retrievers struggle to meet the nuanced requirements of users in real-world applications, particularly in specialized fields like law~\cite{lecardv1, coliee-2023}, healthcare~\cite{trec-covid, mm-ir, rag-healthcare}, and scientific research~\cite{specter, doris, rag-sci}, where precise and context-aware retrieval is crucial~\cite{bio-precise-search, sci-precise-search, weller2025rank1testtimecomputereranking}. 
For instance, in legal research, lawyers often search for target cases using detailed instructions that incorporate specific legal criteria, contextual information, and desired outcomes, as illustrated in \autoref{fig:complex-instruction}. 
However, traditional IR systems~\cite{mimic, liu2021conversational, rq-rag} lack the ability to fully understand and process such complex user instructions in an \emph{end-to-end} manner. 
Consequently, users must break down their complex information-seeking needs into several simpler search queries and manually filter the retrieved cases~\cite{liu2021conversational}, which is time-consuming and inefficient. 
 
On the other hand, existing instruction-following IR benchmarks typically employ simplified instructions, such as single sentence~\cite{su2023one} or a set of keywords~\cite{zhao2024beyond}. 
This simplification in evaluation leads to an incomplete assessment of the model's real-world performance.
Although concurrent works like \followir~\cite{weller2024followir, weller2025mfollowir}, \instructir~\cite{oh2024instructir} and \bright~\cite{BRIGHT} incorporate more complex instructions, they do not establish explicit complexity levels to evaluate retrievers' \emph{fine-grained} abilities in following instructions.
Moreover, these studies primarily focus on general domains, leaving the evaluation of instruction-following retrieval in \emph{expert domains} largely underexplored.

In this work, we introduce \textbf{\ours}, a comprehensive benchmark designed to evaluate the \textbf{\underline{I}}nstruction- \textbf{\underline{\textsc{f}}}ollowing capabilities of  \textbf{\underline{I}}nformation  \textbf{\underline{R}}etrievers, particularly in the context of specialized domains. 
\ours includes eight subsets covering four specialized domains: finance, scientific literature, \lawfield, and healthcare. 
To provide a more granular evaluation, we create three levels of instruction complexity for each domain, representing a range of real-world information retrieval scenarios in expert domains.
\ours includes \nexample instruction-following queries, each averaging 6.14 ground-truth passages. 
To ensure the dataset's high quality, we conduct a comprehensive human expert validation during its construction. 
Moreover, recognizing the limitations of traditional evaluation methods in measuring instruction-following IR performance, we introduce a novel LLM-based metric, \ourmetric, designed to more accurately assess how well retrievers follow instructions.

Through extensive experiments on \nmodel frontier retrievers, including those based on LLMs, we derive three key findings: 
(1) BM25 performs relatively well because the instructions in expert domains contain more glossary terms.
(2) Instruction-tuned retrievers like \instructor~\cite{instructor} do not perform significantly better than their base models, \ie GTRs~\cite{gtr}. This demonstrates that current instruction-tuned retrievers may not be suitable for complex instructions.
(3) Most evaluated models experience performance declines as the complexity of the instructions increases.
(4) LLM-based retrievers demonstrate more robust performance on both nDCG and \ourmetric, highlighting their potential in managing more complex retrieval tasks in specialized domains.

We conclude our main contributions as follows:
\begin{itemize} [leftmargin=*]
\itemsep0em 
\item We introduce \ours, a comprehensive IR benchmark to evaluate the instruction-following ability of information retrievers across specialized domains, meeting their specific demands. The experiments provide insights into \emph{end-to-end} retrieval in specialized domains.
\item We propose \ourmetric, the first LLM-based evaluation method to measure the instruction-following ability of information retrievers.
\item We conduct extensive experiments encompassing a wide range of retrievers, deriving key findings about their instruction-following abilities. Our experimental results reveal the potential of LLM-based retrievers in instruction-following retrieval. 

\end{itemize}
\begin{figure*}[ht]
    \centering
    \includegraphics[width=\textwidth]{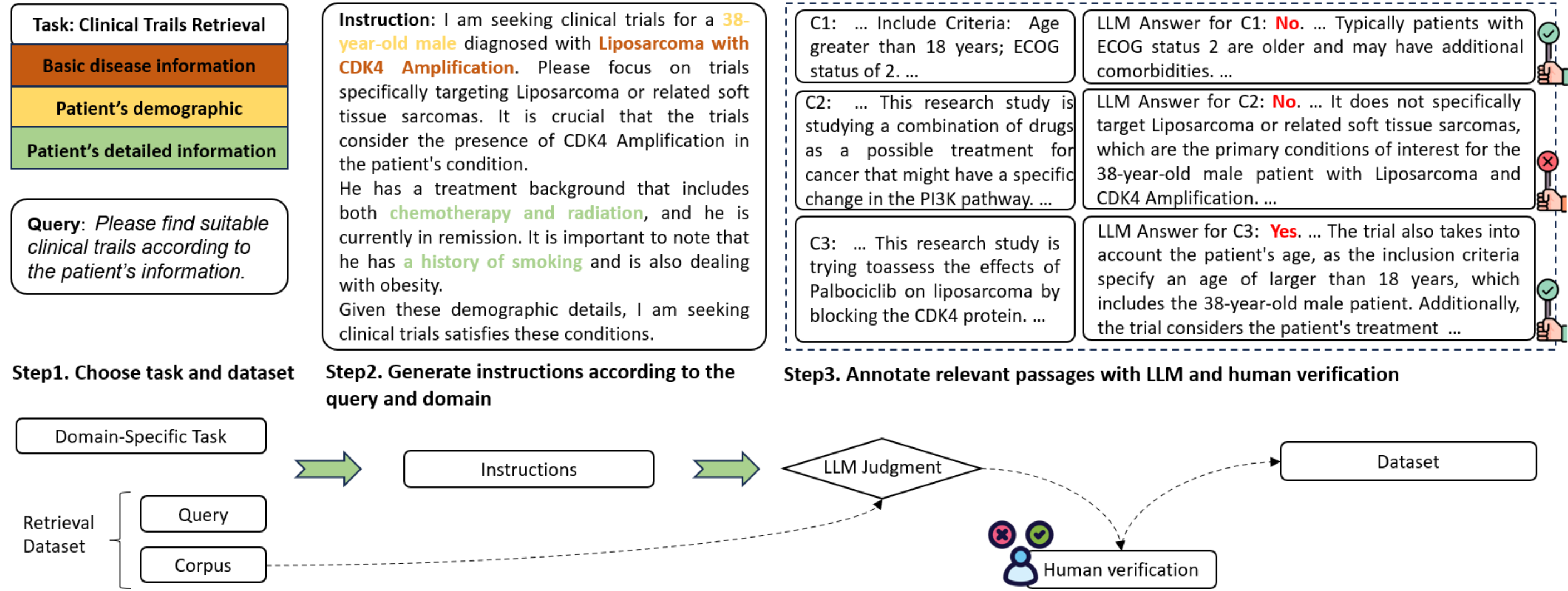}
    \caption{Dataset Construction Pipeline: We derive a specific task according to the dataset, which then guides the generation of instructions based on the original query and task conditions. An LLM is used to assess whether the corpora are relevant to these instructions. As illustrated in the figure, different colors in the ``Task'' section correspond to the conditions outlined in the ``Instruction'' section. }
    \label{fig:data-construction}
\end{figure*}
\section{Related Work}
\subsection{Expert-domain IR Benchmarks}
Information retrieval plays a crucial role in expert domains by enabling efficient access to domain-specific knowledge, facilitating evidence-based decision-making, and accelerating research. 
In recent years, there has been a growing emphasis on developing IR benchmarks in specialized domains, such as \lawfield~\cite{CAIL2019SCMAD, li2023lecardv2}, finance~\cite{fiqa,chen2021finqa, zhao-etal-2024-knowledgefmath}, scientific literature~\cite{scimmir, litsearch}, and healthcare~\cite{trec-covid, 2.1-healthcare-survey,  2.1-healthcare-rag}.
However, existing benchmarks primarily employ oversimplified queries, lacking the depth and specificity required in real-world specialized domains.
For instance, in healthcare, practitioners often formulate complex, context-rich queries that integrate patient-specific information (\eg medical history and treatment plans) to retrieve relevant clinical passages. 
Such queries go beyond simple keyword search and require models to handle complex, user-customized needs. 
This gap underscores the need for a new benchmark that thoroughly evaluates the ability of retrievers to handle instruction-following IR tasks in specialized domains.

\subsection{Instruction-Following IR} 
Several studies have proposed novel training techniques to improve retrievers' instruction-following capabilities~\cite{su2023one, asai2023task, wang2023self}. 
However, due to the lack of instruction-specific IR benchmarks during model development, these approaches have typically been evaluated using traditional benchmarks like BEIR~\cite{beir} and MTEB~\cite{mteb}, which contain queries without complex instructions. 
To address this gap, new instruction-following IR benchmarks have been proposed. 
Specifically, \instructir~\cite{oh2024instructir} reformulates queries in existing retrieval datasets by incorporating user-aligned instruction. However, it limits each query to a single relevant passage, which does not reflect the complexity of real-world scenarios where multiple relevant passages may exist.
While \followir~\cite{weller2024followir} introduces more complex instructions and passage setups, it focuses on the reranking task. 
Moreover, these studies primarily focus on instruction-following retrieval in general domains, leaving the evaluation of instruction-following IR in \emph{expert domains} largely underexplored. 
\section{\ours Benchmark}
We introduce \ours, a comprehensive benchmark designed to assess and enhance retrievers' instruction-following capabilities in expert domains. 
The construction process of \ours, illustrated in \autoref{fig:data-construction}, employs a semi-automated, human-in-the-loop pipeline that ensures both scalability and high quality.
Specifically, we expand the queries in existing specialized-domain IR benchmark (detailed in \S\ref{sec:corpus_collection}) by incorporating detailed instructions that closely mirror real-world scenarios in relevant expert domains (\S\ref{sec:instruction_annotation}). 
Each example is thoroughly validated by domain experts (details of annotator selection and assigned tasks are provided in \autoref{tab:human-valid-detail} in Appendix), ensuring that the instructions are contextually relevant and represent real-world challenges (\S\ref{sec:instruction_annotation}). 
Additionally, the corresponding relevant passages are carefully verified for completeness and accuracy (\S\ref{sec:passage_annotation}).
In the following subsections, we provide a detailed description of each step in the construction pipeline.

\begin{table*}[!t]
\centering
\resizebox{\linewidth}{!}{%
    \begin{tabular}{llrrrp{8cm}}
        \toprule
        \textbf{Domain} & \textbf{Adopted Datasets} & \textbf{\# Qry} & \textbf{Corpus Size} &  \textbf{\# RP} & \textbf{Designed Tasks Reflecting Real-world Challenges}  \\
        \midrule
        Finance 
        & FiQA~\cite{fiqa}  & 1,718 & 57,638 & 3.54 & 
        Retrieve financial suggestions based on user-specific needs to support informed decision-making. \\

        \noalign{\vskip 0.5ex}\hdashline\noalign{\vskip 0.5ex}
        
        Scientific 
        & SciFact-open~\cite{wadden2022scifact_open} & 152 & 500,000 & 4.84 & \multirow{2}{8cm}{Retrieve relevant scientific literature tailored to specific scientific research needs.} \\
        Literature & NFCorpus~\cite{boteva2016nfcorpus} & 86 & 3,633 & 2.81 & \\

        \noalign{\vskip 0.5ex}\hdashline\noalign{\vskip 0.5ex}
        
        \multirow{2}{*}{Law} 
        & AILA~\cite{bhattacharya2019fire}  & 85 & 2,914 & 2.01 & Retrieve legal cases that satisfy customized demands. \\
        & FIRE~\cite{mandal2017overview}  & 168 & 1,745 & 3.36 & Retrieve legal cases to support the judicial decision. \\

        \noalign{\vskip 0.5ex}\hdashline\noalign{\vskip 0.5ex}
        
        \multirow{2}{*}{Healthcare} 
        & TREC-PM~\cite{roberts2017overview, roberts2018overview}  & 172 & 241,006 & 15.61 & 
        \multirow{2}{8cm}{Retrieve relevant clinical trials based on patient's demographics (\eg age, gender, medical history)} \\
        & TREC-CDS~\cite{roberts2015overview} & 43 & 633,955 & 10.84 & \\
        \bottomrule
    \end{tabular}
}

\caption{Overview of the adopted datasets and designed instruction-following IR tasks in \ours, along with basic statistics for each domain. ``\textbf{\# RP}'' represents the average relevant passage number per instruction-following query.
}
\label{tab: Tasks for Logical IR dataset}
\end{table*}
\subsection{Retrieval Dataset Collection}\label{sec:corpus_collection}
To facilitate comprehensive evaluations of instruction-following capabilities across diverse expert domains, \ours spans four specialized domains: finance, scientific literature, law, and healthcare. 
These domains are chosen for their significant demand for precise information retrieval and the complexity of their often nuanced queries. 
For each domain, we extend one or two well-established traditional IR benchmarks (see \autoref{tab: Tasks for Logical IR dataset}), adapting them to instruction-following IR tasks. This ensures that \ours captures the real-world challenges for each expert domain.

\subsection{Instruction Annotation} 
\label{sec:instruction_annotation}
We now describe the process of augmenting the original IR benchmark queries with instructions that mirror real-world demands and challenges specific to each domain. Details of instruction generation are shown in Appendix ~\ref{sec: instruction generation}. 
These instructions introduce complex, often implicit conditions that significantly increase the difficulty of retrieval tasks, as they require the model to not only identify relevant passages but also interpret and follow specific instructions tied to nuanced, task-specific information retrieval.
To ensure high-quality instruction annotation, we employ a \emph{human-in-the-loop} pipeline. We first use LLMs (\ie GPT-4o \texttt{2024-05-13} version) to generate instructions at varying levels of complexity for each query. These instructions are then reviewed and refined by domain experts to ensure they are precise and aligned with the real-world demands of the respective domains.
We detail the annotation process for each domain as follows:

\paragraph{Finance} 
In the finance domain, we focus on the instruction-following IR task centered around personal finance inquiries, simulating scenarios where users seek guidance in making informed financial decisions.
We design three complexity levels of instructions in queries: 
The first level involves simple instructions, \eg ``Please help me to find a financial suggestion for the query \texttt\{\$query\}.''  
The second level includes additional personal information such as age, occupation, and financial status. 
The third level builds upon the second level by incorporating specific financial goals. For example, ``As a 40-year-old accountant with a steady income and moderate savings, I am seeking advice on the best business structure for taxes when combining full-time work with running a small side business. I am looking for insights on how to optimize tax efficiency while balancing the demands of my full-time job and side business.''  

\paragraph{Scientific Literature} In the science literature field, we focus on the science passage searching queries, simulating a person working in a relevant area(\eg teacher, student, etc.) trying to find passages related to scientific claims or problems. We recognize that query instructions can vary in research topics (\eg society, history, biomedical, etc.) and research objectives (\eg influence, reasoning process). We use the Scifact-open dataset to generate three different levels of instructions. The first level of instruction might state, ``Please help me find relevant evidence to support the scientific claim.'' The second level uses previously annotated ``SUPPORT'' and ``CONTRADICT'' tags to generate instructions like ``Please help me to find supporting evidence for this scientific claim.'' The third level includes more customized requirements like research topics and objectives.

\paragraph{Law} In the legal field, we focus primarily on the legal case retrieval task, simulating someone(\ie a lawyer) attempting to find relevant references from previous legal cases. We have two types of instructions. One type is to retrieve prior cases that support the reasoning process for the current case, which originates from the FIRE2017 dataset. The other, derived from AILA2019, is designed to retrieve similar cases according to the demands of legal professionals. For the first type, we construct instructions based on the context in the passage that requires citation. For the second type, we construct three different levels of instructions. The first level, similar to previous domains, is ``Please help me to find cases similar to the current legal case.'' The second level adds conditions including whether the case is beneficial to the defendant or plaintiff. The third level constructs instructions searching for cases relevant to some details of the current case while still satisfying the previous two levels.

\paragraph{Healthcare} In the healthcare domain, we focus on retrieving healthcare-relevant passages (\eg, clinical trials, diagnoses), simulating a doctor finding passages suitable for patients. The seed datasets are derived from TREC-CDS and TREC-PM. Given the two different datasets and corresponding tasks in the biomedical field, TREC-CDS provides a summary accompanied by a detailed description, which we directly use as the instruction. Inspired by the TREC-CDS track, we expand the basic information provided in the TREC-PM track and construct three levels of instructions. The first level contains conditions of the patient's disease and gene variation. The second level adds conditions about the patient's demographics, including age and gender. The third level allows the LLM to create information about the patient's treatment history and family medical history. 

\autoref{tab: Tasks for Logical IR dataset} presents the data statistics of \ours. 
It includes a total of \nexample instruction-following queries across four expert domains. Examples of queries for each domain are provided in Appendix~\ref{inst-examples}.
Each query is carefully reviewed by one of domain experts (\autoref{tab:human-valid-detail} in Appendix).

\subsection{Relevant Passage Annotation}\label{sec:passage_annotation}
We next discuss the process of annotating relevant passages. 
For each instruction-following query in \ours, we select relevant passages from its original dataset. 
The key insight is that if a passage is annotated as relevant to a query, it may also be relevant to an instruction constructed based on that query. Therefore, we need to verify whether these query-relevant passages satisfy the conditions outlined in the corresponding instructions.
Specifically, we first use LLM (\ie GPT-4o) to assess the relevance of each original relevant passage to the instruction. 
The LLM is tasked with generating justification explanations alongside its relevance assessments (we present prompt in Appendix~\ref{exp-settings}).
Human annotators then review the relevance of each passage and the justifications provided by the LLM. If a passage is found to be misaligned with the instruction, the annotators exclude it.

\begin{table}[!t]
\centering
\small
\begin{tabular}{lc}
    \toprule
     \textbf{Evaluation Criteria} & \textbf{Score (1-5)} \\
    \midrule
    \emph{Instruction-Following Query} \\
    \quad Naturalness & 4.24 \\
    \quad Fluency & 4.81 \\
    \quad Expertise & 4.87 \\
    \midrule
    \emph{Relevant Passage (RP)} \\
    \quad Relevant Passage Agreement & 4.43 \\
    
    \noalign{\vskip 0.5ex}\hdashline\noalign{\vskip 0.5ex}
    
    \emph{Excluded RP} \\
    \quad  Exclusion Agreement & 4.32 \\
    \bottomrule
\end{tabular}
\caption{Human Validation Results. Naturalness of instructions evaluates how well the instructions align with real-world demands. The Relevant Passage Agreement Score refers to human annotators' agreement with the LLM on identifying a golden passage, while the Exclusion Agreement Score reflects human annotators' agreement on excluding a passage. }
\label{tab: tab2-human-validation}
\end{table}

\subsection{Dataset Analysis}\label{sec:human_verification}
\autoref{tab: Tasks for Logical IR dataset} presents the basic statistics of \ours.
We also conduct a final human evaluation of data quality on 50 examples from each domain subset. 
For each domain, an external expert (not involved in the data annotation) assesses the quality of each example, providing ratings across several criteria on a scale of 1 to 5. The results, shown in \autoref{tab: tab2-human-validation}, indicate consistently high scores (> 4) for instruction-following query and relevant passages. These evaluation results indicate the high quality of the benchmark proposed by us, further demonstrating its reliability for assessing instruction-following capabilities in relevant tasks.

\begin{table*}[htbp]

\centering
\resizebox{\textwidth}{!}{%
\addtolength{\tabcolsep}{-0.38em}
    \begin{tabular}{lcccccccccccccccc}
        \toprule[.1em]
          & \multicolumn{2}{c}{FiQA} & \multicolumn{2}{c}{SciFact-open} & \multicolumn{2}{c}{NFCorpus} & \multicolumn{2}{c}{AILA} & \multicolumn{2}{c}{FIRE} & \multicolumn{2}{c}{TREC-PM} & \multicolumn{2}{c}{TREC-CDS} & \multicolumn{2}{c}{\textbf{Average}} \\
         \cmidrule(lr){2-3}  \cmidrule(lr){4-5} \cmidrule(lr){6-7} \cmidrule(lr){8-9} \cmidrule(lr){10-11} \cmidrule(lr){12-13} \cmidrule(lr){14-15} \cmidrule(lr){16-17}
          &nDCG & \ourmetric & nDCG & \ourmetric & nDCG & \ourmetric & nDCG & \ourmetric & nDCG & \ourmetric & nDCG & \ourmetric  & nDCG & \ourmetric & \textbf{nDCG} & \ourmetric \\
        \midrule
        \multicolumn{17}{c}{\textit{\textbf{\large Non-instruction-tuned Models}}}\\ \noalign{\vskip 0.8ex}
        GTR-XL & 0.44 & -4.85 & 0.54 & -2.93 & 0.60 & 7.30 & 0.05 & -0.13 & 0.54 & 0.40 & 0.31 & -0.85 & 0.15 & 7.69 & 0.37 & 0.95 \\ 
        BM25 & 0.25 & 1.10 & 0.49 & -0.40 & 0.43 & 1.46 & 0.10 & 0.02 & 0.55 & 0.03 & 0.47 & 3.43 & 0.07 & -2.14 & 0.34 & 0.50 \\ 
        GTR-Large & 0.39 & -6.47 & 0.50 & 0.05 & 0.51 & -3.37 & 0.07 & -0.36 & 0.49 & -5.21 & 0.28 & -1.46 & 0.09 & 11.38 & 0.33 & -0.78 \\ 
        GTR-Base & 0.33 & -4.62 & 0.47 & -0.62 & 0.47 & -0.08 & 0.05 & \textbf{0.54} & 0.52 & 0.81 & 0.27 & -0.65 & 0.13 & 7.03 & 0.32 & 0.35 \\ 
        Contriever & 0.13 & 0.52 & 0.29 & -8.22 & 0.36 & 0.17 & 0.08 & 0.09 & 0.51 & 2.82 & 0.09 & 1.24 & 0.04 & 2.46 & 0.21 & -0.13 \\ 
        ColBERT & 0.07 & 0.17 & 0.14 & 0.34 & 0.16 & 0.06 & 0.07 & -0.01 & 0.39 & 1.44 & 0.02 & 1.03 & 0.00 & -0.94 & 0.12 & 0.30 \\ 
        \midrule
        \multicolumn{17}{c}{\textit{\textbf{\large Instruction-tuned Models}}}\\ \noalign{\vskip 0.8ex}
        NV-Embed-v2 & \textbf{0.68} & 2.76 & \textbf{0.65} & -1.10 & \textbf{0.71} & 13.70 & 0.07 & -0.35 & 0.54 & 0.60 & 0.54 & 0.72 & 0.40 & -5.19 & \textbf{0.51} & 1.59 \\ 
        GritLM-7B & 0.63 & 3.09 & 0.63 & -0.06 & 0.70 & 15.10 & 0.10 & -0.32 & 0.51 & 4.01 & \textbf{0.57} & -0.09 & \textbf{0.42} & -0.32 & \textbf{0.51} & 3.06 \\
         E5-mistral-7B & 0.54 & 4.26 & 0.63 & 0.05 & 0.69 & 14.14 & 0.10 & 0.08 & \textbf{0.57} & \textbf{6.31} & 0.56 & 0.92 & 0.28 & -4.42 & 0.48 & 3.05 \\ 
        Instructor-XL & 0.48 & 1.03 & 0.49 & -2.36 & 0.53 & 0.35 & 0.07 & -0.30 & 0.53 & 1.96 & 0.17 & -2.06 & 0.19 & 0.18 & 0.35 & -0.17 \\ 
        Instructor-Large & 0.49 & 3.65 & 0.46 & 0.20 & 0.56 & 3.68 & 0.07 & 0.29 & 0.51 & 2.19 & 0.15 & -3.86 & 0.17 & 6.70 & 0.34 & 1.84 \\ 
        Promptriever-7B & 0.22 & \textbf{8.95} & 0.34 & \textbf{3.69} & 0.60 & \textbf{18.17} & 0.09 & -0.31 & 0.52 & 5.18 & 0.35 & \textbf{13.26} & 0.09 & 7.07 & 0.32 & \textbf{8.00} \\ 
        Instructor-Base & 0.39 & 3.31 & 0.45 & 0.42 & 0.48 & 2.06 & 0.06 & 0.18 & 0.51 & -2.70 & 0.17 & 1.34 & 0.09 & \textbf{13.93} & 0.31 & 2.65 \\
        \midrule
        \multicolumn{17}{c}{\textit{\textbf{\large Proprietary Models}}}\\ \noalign{\vskip 0.8ex}
        OpenAI-v3-large & 0.54 & 1.57 & 0.59 & -0.48 & 0.58 & 0.31 & \textbf{0.11} & -0.03 & \textbf{0.57} & -0.03 & 0.52 & 0.18 & 0.30 & -5.72 & 0.46 & -0.60 \\
        OpenAI-v3-small & 0.46 & 2.31 & 0.58 & -0.94 & 0.56 & 0.83 & 0.08 & -0.29 & 0.53 & 3.26 & 0.41 & -1.21 & 0.24 & -0.81 & 0.41 & 0.45 \\ 
        \bottomrule[.1em]
    \end{tabular}
}
\caption{Performance of retrievers on \ours measured by nDCG@20 and \ourmetrick (\%). Fine-grained results(\ie hybrid) can be found at Appendix~\ref{sec: detail-exp-res}. Model performance is ranked based on average results with the nDCG metric.} 
\label{tab: main-results}
\end{table*}
\section{Experiment Setup}
This section outlines the experimental setup of our study. We first introduce the two automated metrics used for evaluating \ours, and then discuss the evaluated retrieval systems. Implementation details can be found in Appendix~\ref{exp-settings}.

\subsection{Evaluation Metrics}
\paragraph{nDCG} We use the widely-adopted IR metric, nDCG~\cite{ndcg}, to evaluate \emph{retrieval performance}. Specifically, given a query with instruction $Q$, golden passages $G$, the retrieved passages $P$ are compared against the $G$ using nDCG to quantify the accuracy and relevance of the retrieval.
While nDCG offers a broad assessment of a model's retrieval capabilities, it does not capture the fine-grained aspects of a model's ability to follow instructions. 

\paragraph{\ourmetric} 
To address the aforementioned limitation,
we introduce a new LLM-based metric, \ourmetric, specifically designed to evaluate \emph{instruction-following capabilities} on \ours. 
The core idea is to assess the improvement a retriever demonstrates when instructions are incorporated into the query, compared to when they are not.
We adopt the evaluation prompt from G-Eval~\cite{geval}, as shown in \autoref{fig: aila-eval} in the Appendix, and use GPT-4o-mini as the base evaluator to assess the alignment between each retrieved passage and the given instruction. Considering the potential biases and inaccuracies introduced by LLMs, we use the Probability Normalization technique (as detailed in Appendix~\ref{exp-settings} ) to reduce overestimation, which has been proven effective in current works~\cite{geval, llm-comparative-assessment}. 
For each passage in the Top-$K$ retrieval results, the LLM is instructed to produce the relevance score. 
We then average the scores as the final matching score \(S\). 
The \ourmetric@$K$ for the Top-$K$ retrieved passages is then calculated as:

\begin{equation}
    \ourmetric\text{@K} = (S_\text{inst} - S_\text{q}) \cdot \alpha
    \label{eq:instfol}
\end{equation}

\noindent where \(S_\text{q}\) measures how well the passages retrieved by the original query (without instruction) align with the given instruction; \(S_\text{inst}\) measures how well the passages retrieved by the query (with instruction) meet the same instruction. The factor \(\alpha\) is a normalization function that ensures the \ourmetric score ranges between -1 and 1. In practice, we set \(\alpha\) as \(\frac{1}{3 - S_\text{q}}\).
The implementation details of the \ourmetric metric are provided in Appendix~\ref{exp-settings}.

Our in-depth analysis in Section \ref{sec:metric_analysis} demonstrates that \ourmetric exhibits a high correlation with human expert evaluations, highlighting its reliability. 

\subsection{Evaluated Retrievers}
We evaluate a wide range of retrievers, with model sizes ranging from 110M to 7B parameters. These models are categorized into two main types:

\paragraph{Non-instruction-tuned models}
We include the following commonly-used non-instruction-tuned models for the experiments: (1) \textbf{BM25}~\cite{bm25}, which is a lexical retriever; (2) \textbf{ColBERT}~\cite{colbert}, which encodes queries and documents separately and introduces a mechanism of delayed interaction to be more effective; (3)  \textbf{Contriever}~\cite{contriever}, which is a BERT-based model trained by contrastive learning; and (4) \textbf{GTR}~\cite{gtr}, which uses the encoder from the T5 model and are pre-trained on MS MARCO~\cite{bajaj2018msmarco}. 

\paragraph{Instruction-tuned retrievers}
For the instruction-tuned models, we select: (1) \textbf{\instructor}~\cite{instructor}, which are finetuned on the GTR family using MEDI datasets, and can be utilized for various tasks including retrieval; 
(2) \textbf{E5-mistral-7b-instruct}~\cite{e5}, which is a retriever based on a Mistral model and trained on synthetic data. 
(3) \textbf{GritLM-7B}~\cite{gritlm}, which is also a Mistral model, trained on the synthetic data from E5-mistral-7b-instruct and MEDI2, capable of performing both generation and retrieval tasks;
(4) \textbf{Promptriever}~\cite{promptriever}, is an instruction-trained bi-encoder retriever specialized for instruction-following tasks. 
The model we choose is Promptriever-7B. 
(5) \textbf{NV-Embed}~\cite{nv-embed} is a retriever trained by a two-stage contrastive instruction-tuning method. 
(6) \textbf{Proprietary Retrievers}, including OpenAI's Text-Embedding-v3-Large and Text-Embedding-v3-Small.

\section{Experimental Results}
This section first presents the key experimental results, followed by an in-depth analysis of model performance, an error case study, and a reliability evaluation of the proposed \ours metric.

\subsection{Main Results}
\autoref{tab: main-results} presents the main results, from which we derive the following key findings: 

\paragraph{Non-instruction-tuned models} 
BM25 demonstrates relatively good performance, suggesting possible lexical bias in the datasets. Moreover, GTR models outperform BERT-based models. Unlike ColBERT and Contriever, which are trained solely on the MSMARCO dataset, GTR models also utilize the collected community question-answer pairs and Natural Question~\cite{natural-questions} datasets. These datasets, which are more closely aligned with human interactions, may contribute to the superior performance of GTR models.

\paragraph{Instruction-tuned models} 
(1) The \instructor models show minimal improvement over their backbone (\ie GTR), and in some cases, even perform worse. This may indicate that the \instructor models could be overfitting on specific datasets or are better suited to shorter instructions. 
(2) GritLM-7B, which is of the same size as E5-mistral-7b-instruct, demonstrates stronger performance on healthcare domains where E5-mistral-7b-instruct encounters difficulties. 
This performance gap may stem from GritLM-7B's inclusion of training data in extra domains, which likely boosts its capacity to handle healthcare content more effectively.
(3) Notably, Promptriever-7b outperforms other open-source retrievers in the \ourmetric metric, while NV-Embed-v2 and GritLM-7B demonstrate outstanding performance in the nDCG metric. Promptriever-7b excels in \ourmetric due to its targeted training on instance-level instructions, enabling it to adjust relevance based on user input.
(4) The proprietary retriever, OpenAI-v3-Large, achieves relatively strong performance on nDCG. 
However, both OpenAI's retrievers do not demonstrate superior performance on \ourmetric compared to other retrievers. Unfortunately, the technical details of them, including their training processes, are confidential, which limits our ability to fully understand the factors contributing to their performance. 

\paragraph{Overall} The current training methodologies that integrate instructions are not yet perfect solutions for handling long instructions across various domains. From the relatively good performance of BM25 on both metrics, we can deduce that lexical search may serve as an auxiliary tool for complex instructions in specific domains. 
Meanwhile, although some LLM-based retrievers do not perform well in traditional metrics like nDCG, they exhibit a superior and stable instruction-following ability compared to other retrievers. 
Additionally, the instruction-trained method introduced by Promptriever is highly intuitive and effective, holding promise for future integration into IR systems.

\subsection{Analysis}

\paragraph{Scaling up model size leads to better retrieval performance.} 
From ~\autoref{tab: main-results}, we can conclude that the scaling law applies to retrievers as well. 
Specifically, as model sizes increase from 110M to 1B, both the GTR and \instructor models demonstrate improved nDCG metrics. 
Additionally, LLM-based retrievers(\eg E5-mistral-7b-instruct, GritLM-7B) exhibit relatively strong performance on average. 
However, when considering instruction-following ability, the scaling law does not apply when the model size is below the 1B threshold.
Given the strong performance of GritLM-7B and Promptriever in instruction-following ability, it can be inferred that the current retrieval system can be further enhanced by LLMs finetuned for retrieval tasks.

\paragraph{Existing instruction training methods are still limited.}
Currently, some retrievers, such as \instructor and GritLM-7B, are trained with instructions like ``Retrieve document from Wikipedia'' or ``Classify the question's topic'' to fit the varying demands of different domains. We investigate how significantly such training methods can enhance performance across various domains. 
Accordingly, as described in these works, we incorporate these instructions as prompts in both the query and embedding processes. We format the input as ``\texttt{[Prompt] [Query] [Instruction]}''. The prompt is actually the instruction in these works which gives hints to target tasks and domains, \eg ``Represent the science question for retrieval. '' which is different from our instruction.  We use instruction in these works as a prompt to check whether this is an enhancement compared to embedding with no prompt. 
\begin{figure}[!t]
    \centering
    \includegraphics[width=0.48\textwidth]{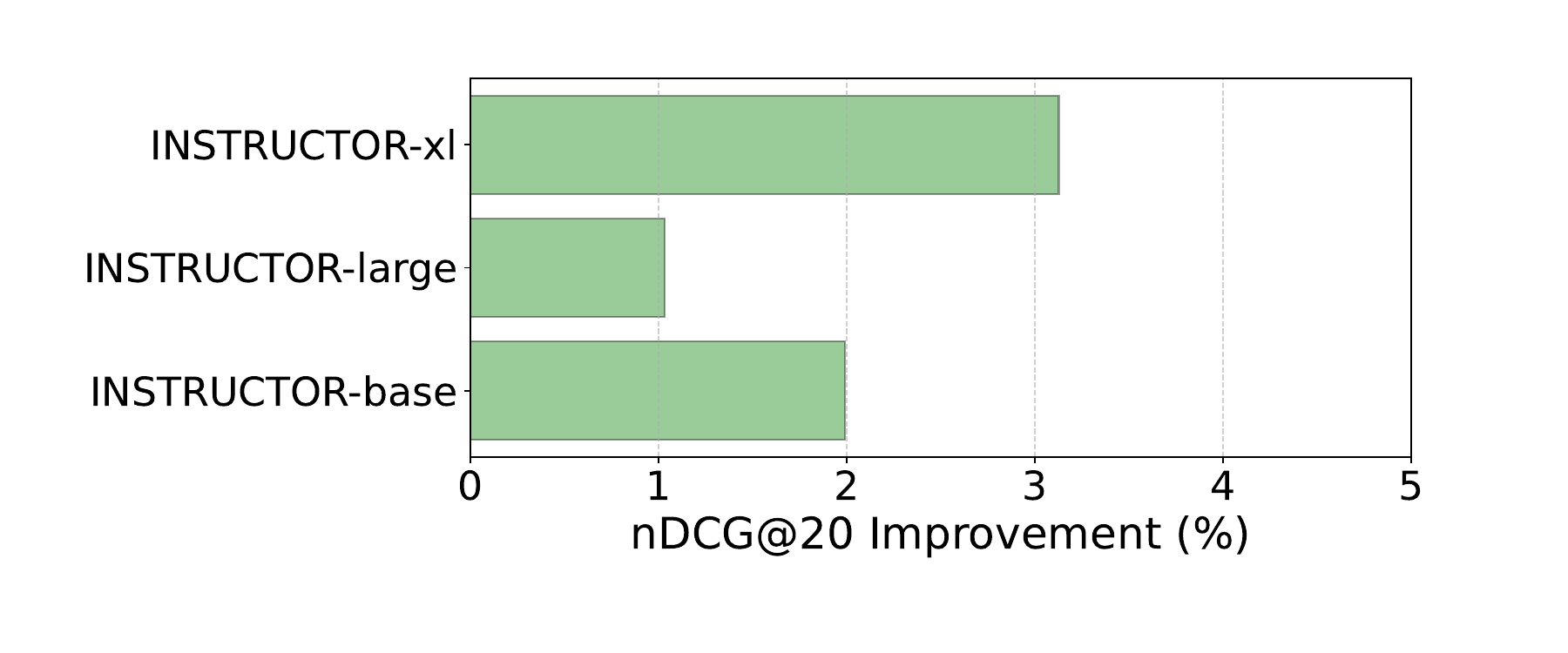}
    \caption{The nDCG improvement when the model is provided with detailed instructions for retrieval.}
    \label{fig: Instruction Improvement}
\end{figure}
The results are shown in ~\autoref{fig: Instruction Improvement}, with detailed outcomes available in the Appendix \ref{sec: ab1}. We observe that adding instructions does not significantly impact the final performance. Additionally, for LLM-based retrievers (\ie GritLM), performance even declines. Therefore, for various domains, merely adding minimal instructions is insufficient. Domain-specific datasets and more complex instructions are required for different domains.

\begin{figure}[!t]
    \centering
    \includegraphics[width=0.45\textwidth]{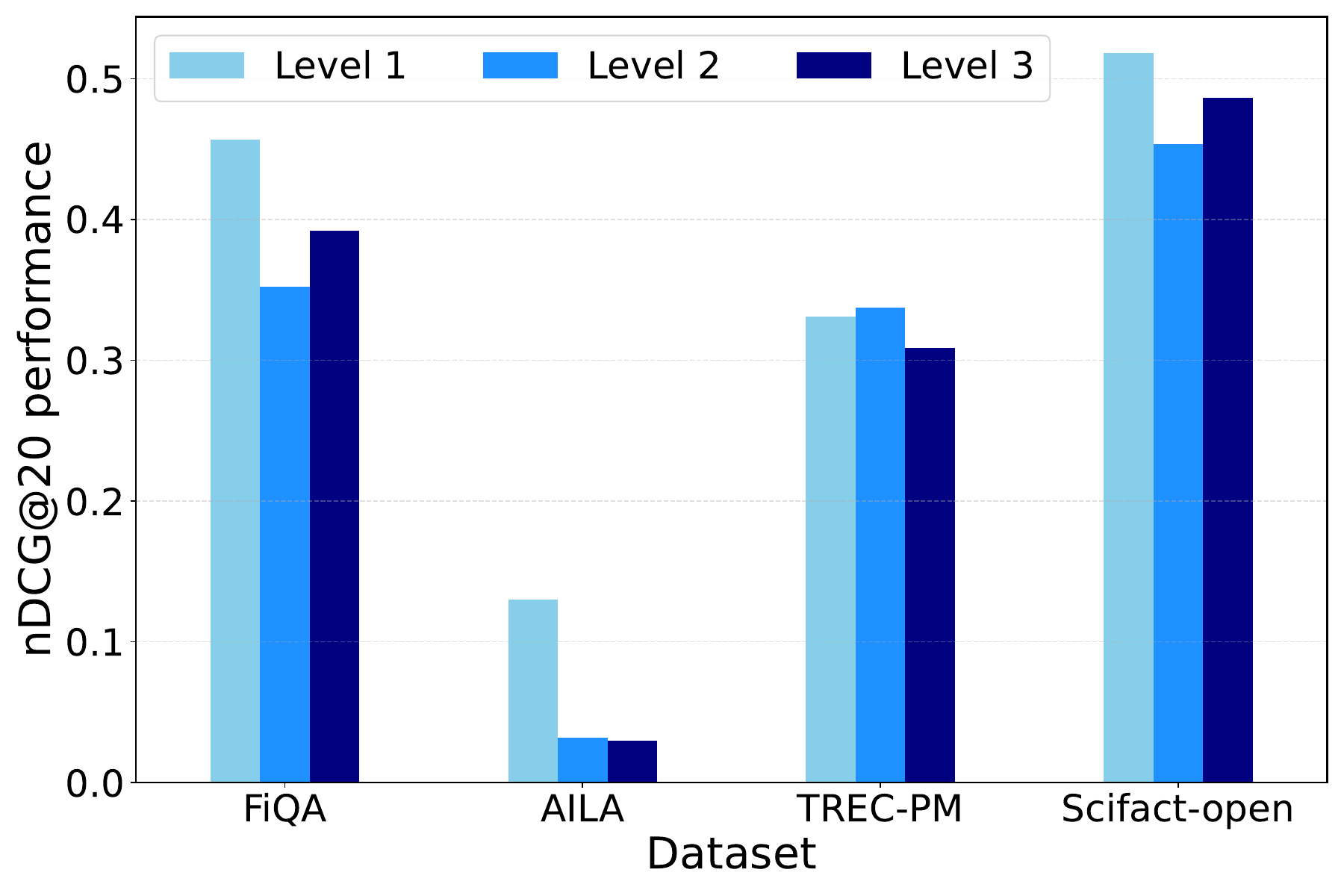}
    \caption{Average nDCG@20 performance on different levels of instructions in different domains.} 
    \label{fig: Levels-comparison}
\end{figure}

\paragraph{Increase in instruction complexity results in performance decline.} 
As discussed in \S\ref{sec:instruction_annotation}, to test the instruction-following abilities of different retrieval systems with finer granularity, we construct instructions with varying complexity levels. 
As shown in ~\autoref{fig: Levels-comparison}, there is a noticeable performance degradation with level2 and level3 instructions compared to level1.
Interestingly, in some subsets, level3 performs better than level2. This improvement is attributed to the fact that level3 instructions are longer and contain more explicit conditions, providing more hints about possible candidates.  
Overall, we observe that retrievers finetuned from LLMs exhibit robust and superior performance compared to other models. This excellent performance may result from LLMs' general capabilities and long-context abilities. Detailed results can be found in the Appendix~\ref{sec: ab2}.

\subsection{Error Analysis}
We select those instructions with both low nDCG scores and \ourmetric scores and create a taxonomy of these instructions, categorizing them as (1) Long Instructions. In the legal domain, some instructions exceed 1,024 tokens due to the inclusion of lengthy legal case. Current retrievers are typically trained with a maximum token length of 512, which cannot perfectly handle these lengthy instructions. (2) Dense with Specialized Knowledge. For instructions that require specialized knowledge, especially in the science literature and healthcare domains, common training data do not cover all the expert knowledge needed in specialized domains. (3) Highly Customized Instructions, as illustrated in Appendix ~\ref{error-analysis-example}. In finance and healthcare domains, users or doctors have several prioritized goals and needs that traditional retrievers may not recognize. 

\section{Reliability Analysis of \ourmetric} \label{sec:metric_analysis}
To evaluate the reliability of the proposed LLM-based evaluation metric, \ourmetric, we conduct a thorough analysis.
Our focus is on \ourmetric's core component: the LLM's ability to score the relevance between an instruction-following query and a passage. 
We validate these LLM-based relevance scores by comparing them with human evaluation scores.
Specifically, we randomly sample 400 query-passage pairs and assign each to one domain expert for evaluation. 
Both the human evaluators and the LLM (\ie GPT-4o) are instructed to rate relevance on a scale of 1 to 5. 
After collecting the scores from both sources, we evaluate their alignment at the instance level. Notably, we observe a strong Pearson correlation coefficient of 0.704, indicating that \ourmetric can effectively assess a retriever's instruction-following capabilities.
\section{Conclusion}
We introduce \ours, a novel IR benchmark designed to assess the instruction-following capabilities of retrievers.  
\ours focuses on domain-specific instructions, reflecting the diverse needs across various fields. 
Our experiments reveal that current instruction-tuned models struggle with long, complex instructions. Moreover, as the complexity increases, a noticeable performance decline occurs across all tested retrieval systems. However, LLM-based retrievers demonstrate more robust performance and relatively better results compared to other models. This suggests potential solutions for end-to-end specialized-domain retrieval scenarios.
\section*{Limitations}
In this section, we list three limitations of this study, each of which opens avenues for future improvements. 
First, our dataset has a limited number of queries accompanied by instructions, and we do not provide a training dataset for future works to train their models. 
Second, we do not compare the performance of domain-specific retrievers like BioBERT~\cite{biobert} with general-purpose retrievers. Domain-specific retrievers, equipped with specialized training data, may achieve superior results in niche fields. Future work should explore the integration and evaluation of such specialized models, particularly in domains like healthcare or law, where domain knowledge is crucial. 
Finally, there are some relevance judgment gaps~\cite{relevance-judge} in the selected seed datasets. Since we select relevant passages based on previously annotated passages, this approach may lead to some relevant passages being ignored. Future work could investigate ways to improve relevance judgment, potentially through the use of more crowd-sourced human annotation.

\bibliography{anthology,custom, llm}

\appendix


\newpage
\clearpage
\section{\ours Benchmark}

\subsection{Instruction Generation}
\label{sec: instruction generation}
We first ask LLM (\ie GPT-4o) to generate an instruction according to the reality demands, with the prompts on dataset NFCoprus shown in ~\autoref{fig: nfcorpus prompt}. And then we check whether previous annotated passages for the query satisfy the generated instruction, with the prompts shown in ~\autoref{fig: nfcorpus evaluate}.

For the datasets with multi-levels, including FiQA, SciFact-open, AILA, and TREC-PM, we do not construct a fully complex instruction all at once. Since we have different levels of reasoning, we ask the LLM to detail the instruction level by level, akin to a bottom-up approach. For example, for a given instruction at level 2, such as ``Please help me to find the plaintiff's beneficial legal case.'' we request the LLM to generate a more complex instruction for the next level that also includes the ``plaintiff's beneficial'' condition. The prompt for instruction generation on dataset AILA is shown in ~\autoref{fig: aila-inst}.

\begin{figure*}[htbp]
\centering
\footnotesize
\fbox{%
    \begin{minipage}{0.9\textwidth}
        \textbf{[system prompt]} \\
        You are an expert in science. \\
        
        \textbf{[user input]} \\
        Given the scientific claim: \{claim\},
Imagine you are a student or researcher seeking information on a specific topic. Based on the conditions listed below, construct a detailed retrieval instruction tailored to the claim. You do not need to incorporate all of the conditions, but ensure your instruction is relevant. \\
    * Research fields \\ 
    * Research topics \\
    * Research objectives \\
    * Customized needs (For example, experimental subjects, experimental methods, etc.) \\

The instruction should target a single type of information and be both coherent and logical. It should also be detailed and specific, presented in the first person, and narrated naturally in one paragraph. \\

Please return your answer as follows: \\
Instruction: ... \\
    \end{minipage}%
}
\caption{Prompt for generating instruction on NFCorpus dataset.}
\label{fig: nfcorpus prompt}
\end{figure*}

\begin{figure*}[htbp]
\centering
\footnotesize
\fbox{%
    \begin{minipage}{0.9\textwidth}
        \textbf{[system prompt]} \\
        You are an expert in science.
        
        \textbf{[user input]} \\
        Given an instruction \{instruction\}, and an corpus \{corpus\}, check whether the instruction is satisfied by the corpus.\\
Please only return 'yes' or 'no' and your reason, and return in the following format. \\
Answer: yes/no
Reason: ...
    \end{minipage}%
}
\caption{Prompt for evaluating corpus on NFCorpus dataset.}
\label{fig: nfcorpus evaluate}
\end{figure*}

\begin{figure*}[htbp]
\footnotesize
\centering
\fbox{%
    \begin{minipage}{0.9\textwidth}
        \textbf{[system prompt]} \\
        You are an expert in the legal domain.
        
        \textbf{[user input]} \\
        Given a legal case: \\
\{case\}. \\ 
And an instruction: \{instruction\}. \\ 

Please provide a detailed instruction based on the case. Include specific situations from the case to elaborate on the instruction. Your response should be narrated as if you are examining various cases, and it should be presented in a single paragraph. \\ 
The instruction should not be longer than 2-3 sentences. \\

For example: \\ 
Legal Case: "XYZ Corporation filed a lawsuit against John Doe, a former employee, for defamation after Doe posted allegations on his personal social media claiming that XYZ provided false information to customers. While Doe's post didn't result in significant or immediate financial loss for the company, XYZ argued that it tarnished their reputation." \\
Instruction: "I'm the plaintiff's lawyer and I'm looking for civil tort cases involving the right to reputation and lowered social evaluation, particularly where an employee posted on social media that the company made false statements in providing services but no serious consequences occurred, and it's difficult to prove the lowered social evaluation." \\
Your response should be formatted as follows: \\
Instruction: ... \\
    \end{minipage}%
}
\caption{Prompt for instruction generation on AILA dataset.}
\label{fig: aila-inst}
\end{figure*}



\begin{table*}[htbp]
    \centering
    \footnotesize
    \scalebox{1.0}{
        \begin{tabular}{lllp{7cm}}
        \toprule[.1em]
            ID & Fluent in English & Major & Annotation tasks \\ 
        \midrule
            1 & > 10 yrs & Legal & Legal subset annotation \\ 
            2 & > 7 yrs & Legal & Legal subset annotation \\ 
            3 & > 10 yrs & Biology & Healthcare subset annotation; Science Literature subset annotation; Annotation Validation \\ 
            4 & > 7 yrs & Pharmacy & Healthcare subset annotation; Science Literature subset annotation \\ 
            5 & > 10 yrs & Biomedical engineering & Healthcare subset annotation; Science Literature subset annotation; Annotation Validation \\ 
            6 & > 7 yrs & Mathematics & Financial subset annotation; Annotation Validation. \\ 
            7 & > 6 yrs & Finance & Financial subset annotation; \\ 
            8 & Native Speaker & Finance & Financial Subset Annotation \\ 
            9 & > 10 yrs & Legal & Legal subset annotation \\ 
        \bottomrule[.1em]
        \end{tabular}
    }
    \caption{Human annotator's tasks}
    \label{tab:human-valid-detail}
\end{table*}

\subsection{Details of Instructions in Each Domain}
\label{inst-examples}
Instruction examples are shown in ~\autoref{tab: Examples of instruction}. For the legal datasets, which belong to the legal domain, the query part consists of only a summary or is omitted due to the length context of legal cases. 

Different complexity instruction examples are shown in ~\autoref{tab: different level}. To emphasize, we describe the content of each level again. As the level increases, so do the conditions. Target passages in Level 3 must satisfy the conditions of Levels 1 and 2, and Level 2 candidates must satisfy Level 1 conditions. 

\noindent(1) \textbf{FiQA}: The first level simply asks for financial suggestions. The second level includes information about personal financial status. The third level incorporates personal financial purposes. 

\noindent(2) \textbf{Scifact-open}: The first level involves asking for science passages relevant to a given science claim. The second level seeks evidence that either contradicts or supports this claim. The third level is tailored for students or researchers who need to find evidence based on customized demands. 

\noindent(3) \textbf{AILA}: The first level involves searching for similar cases. The second level requires that the relevant case be beneficial for the plaintiff or defendant. The third level adds more explicit conditions such as the details of the current cases, potential goals, and more similar scenarios. 

\noindent(4) \textbf{TREC-PM}: The first level includes information about the patient's disease. The second level adds the patient's demographics, including age and gender. The third level incorporates additional information about the patient's treatment history and family history. 

\section{Implementation Details}
\label{exp-settings}
\paragraph{Embedding} To accommodate the long context of certain passages, we employ a sliding window of 512 tokens with an overlap of 128 tokens, using mean pooling to generate embeddings. For LLM-based retrievers, however, we do not apply mean pooling due to their extended context window.
Given hardware constraints, we run LLM-based retrievers in FP16 mode to reduce GPU memory usage. When querying, we concatenate the query and instruction with a space character.
Because of the large number of passages and the correspondingly large embeddings, we use LangChain and Elasticsearch in our experiments. In addition, we also provide the BEIR code. 

\paragraph{Implementation Details of \ourmetric}
Given  $C_{\text{q}}$, which is the retrieved passages set for a query, and $C_{\text{inst}}$, which is the retrieved corpus set for the query combined with instructions, we evaluate each passage in both sets using LLM (\ie \texttt{gpt-4o-mini}) with some evaluation criteria. This approach is inspired by G-Eval~\cite{geval} and TREC's principles of data collection. We then obtain two sets of weighted scores, \(S_q\) and \(S_{\text{inst}}\).  \(S_\text{q}\) measures how well the passages retrieved by the original query (without instruction) align with the given instruction; \(S_\text{inst}\) measures how well the passages retrieved by the query (with instruction) meet the same instruction. 
Each score in \(S_q\) and \(S_{\text{inst}}\) is calculated as follows, where \(s_k\) is an element of \(S_q\) or \(S_{\text{inst}}\), and \(p_k\) represents the logarithmic probability of each score as determined by the API:
\[
\text{weighted\_score} = \frac{\sum_{k=1}^{10} (s_k \times e^{p_k})}{\sum_{k=1}^{10} e^{p_k}}
\]
Here, \( e^{p_k} \) converts the log probabilities back to standard probabilities for calculation purposes. This formula accounts for the inherent probabilistic nature of LLMs, where predictions for each token are based on a statistical probability distribution influenced by configurations such as temperature and top\_p.

We use the average of \(S_\text{q}\) and \(S_{\text{inst}}\) to calculate the instruction-following ability through a metric we propose, called \ourmetric. The insight is to consider the maximum improvement a retriever can achieve. Consider a case with two students, A and B. Student A has a rank of 300 and a previous rank of 500, while student B has a rank of 10 and a previous rank of 40. Traditionally, we would calculate improvement through absolute differences. However, student B has less room to improve his rank. Based on this insight, we propose the \ourmetric metric to evaluate the retriever's instruction-following ability. The factor \(\alpha\) is a normalization function that ensures the \ourmetric score ranges between -1 and 1. In practice, we set \(\alpha\) as \(\frac{1}{3 - S_\text{q}}\).

\begin{equation}
    \ourmetric\text{@K} = (S_\text{inst} - S_\text{q}) \cdot \alpha
    \label{eq:instfol-detail}
\end{equation}

When calling the API to evaluate the \ourmetric, we use top\_p = 0.7 and top\_logprobs = 5. We set the temperature to 0.0 to reduce the overestimation by the LLM. The prompt for evaluation is shown in ~\autoref{fig: aila-eval}.  

\paragraph{Error Analysis}
\label{error-analysis-example}
The example of error analysis is illustrated in ~\autoref{tab: error-analysis}.

\begin{table*}[htbp]
\centering
\scalebox{0.70}{
    \begin{tabular}{cp{7cm}p{12cm}}
        \toprule
        Dataset & query & instruction \\
        \midrule
        FiQA & Full-time work + running small side business: Best business structure for taxes? & As a 40-year-old accountant with a steady income and moderate savings, I am seeking advice on the best business structure for taxes when combining full-time work with running a small side business. I am particularly interested in understanding the tax implications, legal considerations, and potential benefits of different business structures. Additionally, I am looking for insights on how to optimize tax efficiency while balancing the demands of my full-time job and side business. \\
        SciFact-open & A deficiency of folate decreases blood levels of homocysteine. & As an expert in the field of science, I need to find a peer-reviewed research article or a review paper that presents contradicting evidence regarding the relationship between folate deficiency and homocysteine levels in the blood. The passage should offer evidence that opposes the claim stating that a deficiency of folate results in decreased blood levels of homocysteine. \\
        NFCorpus & Why are Cancer Rates so Low in India? &  I am a student researching the factors contributing to low cancer rates in India, and I am specifically interested in understanding the role of dietary habits. I need to find scientific studies or articles from the fields of oncology, nutrition, and epidemiology that focus on the relationship between Indian dietary patterns and cancer prevention. My objective is to analyze the types of foods commonly consumed in India and their potential protective effects against cancer. To meet my customized needs, I require information on specific dietary components, such as spices, fruits, vegetables, and traditional Indian dishes, that have been associated with lower cancer rates. Additionally, I am interested in any experimental studies or clinical trials investigating the effects of these dietary factors on cancer cells or animal models.  \\
        AILA &  The appellant, once a prime witness in a bribery trial, became a Cabinet Minister and resigned after critical judicial remarks during an appeal that acquitted the first respondent. The High Court questioned the evidence and the appellant’s credibility, overturning the initial conviction for accepting bribes. & I represent the appellant and I seek cases involving a defendant who benefitted from a reversal of a conviction due to lack of acceptable evidence and a plausible explanation for the incriminating evidence found in their possession, despite adverse remarks made by the Appellate Judge regarding the credibility of the appellant's testimony in a bribery case where the defendant was acquitted based on insufficient prosecution evidence. \\
        FIRE & [A legal case summary] What was the decision and legal principle established in the case referred to as [?CITATION?] in relation to the doctrine of promissory estoppel in the context of government representations and obligations? & Retrieve the prior case referred to as [?CITATION?] and focus on the court's analysis and ruling regarding the application of promissory estoppel against the government, particularly in situations where representations are made by governmental authorities and the subsequent obligations arising from such representations. Pay attention to any discussion on the enforceability of promises made by the government, the limitations of promissory estoppel against the government, and the factors determining the applicability of the doctrine in cases involving governmental representations. \\
        TREC-PM & A patient diagnosed with Liposarcoma with CDK4 Amplification. I am looking for possible clinical trials suitable for this patient. & I am seeking clinical trials for a 38-year-old male diagnosed with Liposarcoma with CDK4 Amplification. Please focus on trials specifically targeting Liposarcoma or related soft tissue sarcomas. It is crucial that the trials consider the presence of CDK4 Amplification in the patient's condition. Additionally, the patient's age and gender should be taken into account when selecting suitable clinical trial options. Patient Profile: The patient is a 38-year-old male who has been diagnosed with Liposarcoma with CDK4 Amplification. He has a treatment background that includes both chemotherapy and radiation, and he is currently in remission. It is important to note that he has a history of smoking and is also dealing with obesity. Given these demographic details, I am seeking clinical trials that specifically target Liposarcoma or related soft tissue sarcomas, taking into consideration the presence of CDK4 Amplification. The trials should also consider the patient's age and gender, as well as any potential influences from his treatment background, smoking history, and obesity.  \\
        TREC-CDS & Given some infomation about patient. 58-year-old woman with hypertension and obesity presents with exercise-related episodic chest pain radiating to the back.What is the patient's diagnosis? & A 58-year-old African-American woman presents to the ER with episodic pressing/burning anterior chest pain that began two days earlier for the first time in her life. The pain started while she was walking, radiates to the back, and is accompanied by nausea, diaphoresis and mild dyspnea, but is not increased on inspiration. The latest episode of pain ended half an hour prior to her arrival. She is known to have hypertension and obesity. She denies smoking, diabetes, hypercholesterolemia, or a family history of heart disease. She currently takes no medications. Physical examination is normal. The EKG shows nonspecific changes. \\
        \bottomrule
    \end{tabular}
}

\caption{Examples of instructions in different domains.}
\label{tab: Examples of instruction}
\end{table*}

\begin{table*}[htbp]
\centering
\scalebox{0.60}{
    \begin{tabular}{cp{4cm}p{6cm}p{12cm}}
        \toprule
        Dataset & level1 & level2 & level3 \\
        \midrule
        FiQA & Please help me to find the financial suggestions for my query. & I am a 40-year-old accountant with a steady income and moderate savings. & As a 40-year-old accountant with a steady income and moderate savings, I am seeking advice on the best business structure for taxes when combining full-time work with running a small side business. I am particularly interested in understanding the tax implications, legal considerations, and potential benefits of different business structures. Additionally, I am looking for insights on how to optimize tax efficiency while balancing the demands of my full-time job and side business \\
        SciFact-open & Please find the science passage which related to the claim & Please help me to find the contradict evidence. & As an expert in the field of science, I need to find a peer-reviewed research article or a review paper that presents contradicting evidence regarding the relationship between folate deficiency and homocysteine levels in the blood. The passage should offer evidence that opposes the claim stating that a deficiency of folate results in decreased blood levels of homocysteine. \\
        AILA & Please help me find the relevant legal cases. & As a defendant player, I want the case where the defendant is beneficial. & I represent the appellant and I seek cases involving a defendant who benefitted from a reversal of a conviction due to lack of acceptable evidence and a plausible explanation for the incriminating evidence found in their possession, despite adverse remarks made by the Appellate Judge regarding the credibility of the appellant's testimony in a bribery case where the defendant was acquitted based on insufficient prosecution evidence.  \\
        TREC-PM & I'm looking for clinical trials suitable for a 38-year-old male patient diagnosed with Liposarcoma with CDK4 Amplification. & I am seeking clinical trials for a 38-year-old male diagnosed with Liposarcoma with CDK4 Amplification. Please focus on trials specifically targeting Liposarcoma or related soft tissue sarcomas. It is crucial that the trials consider the presence of CDK4 Amplification in the patient's condition. Additionally, the patient's age and gender should be taken into account when selecting suitable clinical trial options. & I am seeking clinical trials for a 38-year-old male diagnosed with Liposarcoma with CDK4 Amplification. Please focus on trials specifically targeting Liposarcoma or related soft tissue sarcomas. It is crucial that the trials consider the presence of CDK4 Amplification in the patient's condition. Additionally, the patient's age and gender should be taken into account when selecting suitable clinical trial options. Patient Profile: The patient is a 38-year-old male who has been diagnosed with Liposarcoma with CDK4 Amplification. He has a treatment background that includes both chemotherapy and radiation, and he is currently in remission. It is important to note that he has a history of smoking and is also dealing with obesity. Given these demographic details, I am seeking clinical trials that specifically target Liposarcoma or related soft tissue sarcomas, taking into consideration the presence of CDK4 Amplification. The trials should also consider the patient's age and gender, as well as any potential influences from his treatment background, smoking history, and obesity. \\
        \bottomrule
    \end{tabular}
}

\caption{Examples for different levels' instruction in various domains.}
\label{tab: different level}
\end{table*}

\begin{table*}[htbp]
\centering
\scalebox{0.7}{
    \begin{tabular}{p{3cm}p{18cm}}
        \toprule[.1em]
        Type & Example \\
        \midrule[.1em]
        Long Instruction & [A long legal case] As the defendant player, seek cases where the prosecution's evidence relies heavily on circumstantial evidence and lacks direct proof of intent or direct involvement in the alleged crime, similar to a situation where the accused individuals were convicted based on circumstantial evidence and witness testimonies, despite maintaining their innocence throughout the trial and appeal process. \\
        Dense with specialized knowledge &  CHEK2 has a significant role in breast cancer As a scientist investigating the claim that 'CHEK2 has a significant role in breast cancer,' I should search for research articles or review papers that provide support evidence on the specific functions of the CHEK2 gene in relation to breast cancer development. \\
        Highly customized instructions & I am seeking clinical trials suitable for a 35-year-old female diagnosed with colorectal cancer and exhibiting FGFR1 Amplification. Please prioritize trials that focus on colorectal cancer specifically or a narrower focus related to this patient's condition. Additionally, it is crucial to include trials that directly match the FGFR1 Amplification gene mutation in the patient. The patient's age and gender are also important factors to consider in selecting appropriate clinical trials. Please ensure that the trials selected meet these criteria for optimal patient care and treatment options. \\
        \bottomrule[.1em]
    \end{tabular}
}

\caption{Taxonomy of instructions with low nDCG score and \ourmetric score. }
\label{tab: error-analysis}
\end{table*}
\begin{figure*}[htbp]
\centering
\footnotesize
\fbox{%
    \begin{minipage}{0.97\textwidth}
        \textbf{[system prompt]} \\
        You are an expert in legal domain. \\
        
        \textbf{[user input]} \\
        Given an instruction: {instruction}, \\
and a prior case: {corpus}, \\ 
please evaluate the prior case according to the instruction and Evaluation Criteria and return a JSON object with the score and reason. \\ 

There are 3 relevant levels to evaluate the case regarding the instruction: \\ 
1. The prior case is similar to the one in the instruction. \\ 
2. The prior case satisfies the instruction at the 'plaintiff' or 'defendant' beneficial level. \\ 
3. The prior case totally matches the instruction, including the detailed requirements in the instruction. \\ 

Evaluation Criteria: \\
1. If the prior case only meets the instruction at the first level, the score is 1. \\ 
2. If the prior case meets the instruction at the first and second levels, the score is 2. \\ 
3. If the prior case meets the instruction at all three levels, the score is 3. \\ 
4. If the prior case does not meet any of the levels, the score is 0. \\ 

Please give a score between 0 and 3. \\ 

** \\ 
IMPORTANT: Please make sure to only return in JSON format, with the "score" and "reason" key. No additional words or explanations are needed. \\ 
Please think step by step about the reason and give the score according to the Evaluation Criteria. \\

Example JSON: \\
{{ \\ 
    "score": 1, \\ 
    "reason": "The corpus only matches the instruction in terms of research field and research topics." \\
}} \\
**\\

JSON:\\
    \end{minipage}%
}
\caption{Prompt for instruction generation on AILA dataset.}
\label{fig: aila-eval}
\end{figure*}

\section{Details of Experimental Results}
\label{sec: detail-exp-res}

\subsection{Detailed Results of \ours}
In addition to the results obtained using \ourmetric, we also present the outcomes of traditional IR metrics, as shown in ~\autoref{tab: main-results-wo-instfol}. Furthermore, beyond the \emph{end-to-end} retrieval method, we implement a \emph{hybrid} retrieval approach utilizing BM25. An overall improvement in performance can be observed for GTR-xl, Instructor-xl, and Promptriever compared to the \emph{end-to-end} method. 
However, as shown in ~\autoref{tab: main-results-wo-instfol}, performance varies across tasks. This highlights the importance of designing a carefully tailored hybrid retrieval pipeline to meet the demands of real-world scenarios.

\begin{table*}[htbp]

\centering
\resizebox{0.96\textwidth}{!}{%
\addtolength{\tabcolsep}{-0.2em}
    \begin{tabular}{lrrrrrrrrrrrrrrrr}
        \toprule[.1em]
          & \multicolumn{2}{c}{FiQA} & \multicolumn{2}{c}{SciFact-open} & \multicolumn{2}{c}{NFCorpus} & \multicolumn{2}{c}{AILA} & \multicolumn{2}{c}{FIRE} & \multicolumn{2}{c}{TREC-PM} & \multicolumn{2}{c}{TREC-CDS} & \multicolumn{2}{c}{Average} \\
         \cmidrule(lr){2-3}  \cmidrule(lr){4-5} \cmidrule(lr){6-7} \cmidrule(lr){8-9} \cmidrule(lr){10-11} \cmidrule(lr){12-13} \cmidrule(lr){14-15} \cmidrule(lr){16-17}
          &nDCG & MRR & nDCG & MRR & nDCG & MRR & nDCG & MRR & nDCG & MRR & nDCG & MRR  & nDCG & MRR & nDCG & MRR \\
        \midrule
        \multicolumn{16}{c}{\emph{\textbf{End-to-end retrieval }}} \\

        \midrule
        GTR-XL & 0.44 & 0.40 & 0.54 & 0.49 & 0.60 & 0.57 & 0.05 & 0.04 & 0.54 & 0.50 & 0.31 & 0.27 & 0.15 & 0.12 & 0.37 & 0.34 \\
        BM25 & 0.25 & 0.21 & 0.49 & 0.45 & 0.43 & 0.40 & 0.10 & 0.08 & 0.55 & 0.51 & 0.47 & 0.43 & 0.07 & 0.05 & 0.34 & 0.30 \\ 
        GTR-Large & 0.39 & 0.34 & 0.50 & 0.46 & 0.51 & 0.46 & 0.07 & 0.07 & 0.49 & 0.41 & 0.28 & 0.23 & 0.09 & 0.06 & 0.33 & 0.29 \\ 
        GTR-Base & 0.33 & 0.28 & 0.47 & 0.43 & 0.47 & 0.42 & 0.05 & 0.04 & 0.52 & 0.47 & 0.27 & 0.24 & 0.13 & 0.09 & 0.32 & 0.28 \\
        Contriever & 0.13 & 0.10 & 0.29 & 0.24 & 0.36 & 0.29 & 0.08 & 0.06 & 0.51 & 0.48 & 0.09 & 0.06 & 0.04 & 0.04 & 0.21 & 0.18 \\ 
        ColBERT & 0.07 & 0.05 & 0.14 & 0.12 & 0.16 & 0.13 & 0.07 & 0.05 & 0.39 & 0.35 & 0.02 & 0.01 & 0.00 & 0.00 & 0.12 & 0.10 \\ 
        \noalign{\vskip 0.5ex}\hdashline\noalign{\vskip 0.5ex}
        NV-Embed-v2 & \textbf{0.68} & \textbf{0.66} & \textbf{0.65} & \textbf{0.62} & \textbf{0.71} & \textbf{0.70} & 0.07 & 0.04 & 0.54 & 0.51 & 0.54 & 0.53 & 0.40 & 0.36 & \textbf{0.51} & \textbf{0.49} \\ 
        GritLM-7B & 0.63 & 0.61 & 0.63 & 0.60 & 0.70 & 0.69 & 0.10 & 0.06 & 0.51 & 0.46 & \textbf{0.57} & \textbf{0.54} & \textbf{0.42} & \textbf{0.38} & \textbf{0.51} & 0.48 \\ 
         E5-mistral-7b-inst & 0.54 & 0.51 & 0.63 & 0.59 & 0.69 & 0.67 & 0.10 & 0.06 & 0.57 & \textbf{0.54} & 0.56 & 0.52 & 0.28 & 0.23 & 0.48 & 0.45 \\
        Instructor-XL & 0.48 & 0.44 & 0.49 & 0.44 & 0.53 & 0.47 & 0.07 & 0.05 & 0.53 & 0.49 & 0.17 & 0.12 & 0.19 & 0.15 & 0.35 & 0.31 \\ 
        Instructor-Large & 0.49 & 0.45 & 0.46 & 0.42 & 0.56 & 0.51 & 0.07 & 0.06 & 0.51 & 0.45 & 0.15 & 0.11 & 0.17 & 0.12 & 0.34 & 0.30 \\ 
        Promptriever-7B & 0.22 & 0.17 & 0.34 & 0.28 & 0.60 & 0.59 & 0.09 & 0.07 & 0.52 & 0.48 & 0.35 & 0.29 & 0.09 & 0.06 & 0.32 & 0.28 \\ 
        Instructor-Base & 0.39 & 0.34 & 0.45 & 0.42 & 0.48 & 0.42 & 0.06 & 0.05 & 0.51 & 0.46 & 0.17 & 0.13 & 0.09 & 0.07 & 0.31 & 0.27 \\
        \noalign{\vskip 0.5ex}\hdashline\noalign{\vskip 0.5ex}
        OpenAI-v3-large & 0.54 & 0.51 & 0.59 & 0.54 & 0.58 & 0.55 & \textbf{0.11} & \textbf{0.08} & \textbf{0.57} & 0.54 & 0.52 & 0.46 & 0.30 & 0.26 & 0.46 & 0.42 \\
        OpenAI-v3-small & 0.46 & 0.41 & 0.58 & 0.52 & 0.56 & 0.52 & 0.08 & 0.06 & 0.53 & 0.48 & 0.41 & 0.38 & 0.24 & 0.21 & 0.41 & 0.37 \\ 
        
        \midrule
        \multicolumn{16}{c}{\emph{\textbf{Hybrid retrieval }}} \\
        \midrule
        GritLM-7B & \textbf{0.59} & \textbf{0.54} & \textbf{0.62} & \textbf{0.58} & \textbf{0.57} & \textbf{0.51} & \textbf{0.10} & 0.07 & 0.55 & 0.50 & \textbf{0.59} & \textbf{0.56} & \textbf{0.39} & \textbf{0.32} & \textbf{0.49} & \textbf{0.44} \\ 
        GTR-XL & 0.43 & 0.38 & 0.57 & 0.51 & 0.52 & 0.47 & 0.06 & 0.06 & \textbf{0.56} & \textbf{0.53} & 0.33 & 0.29 & 0.14 & 0.13 & 0.37 & 0.34 \\ 
        Instructor-XL & 0.46 & 0.41 & 0.54 & 0.51 & 0.52 & 0.46 & 0.09 & \textbf{0.07} & 0.56 & 0.52 & 0.23 & 0.20 & 0.20 & 0.16 & 0.37 & 0.33 \\ 
        Promptriever-7B & 0.25 & 0.21 & 0.42 & 0.38 & 0.54 & 0.49 & 0.09 & 0.07 & 0.56 & 0.53 & 0.42 & 0.37 & 0.11 & 0.09 & 0.34 & 0.31 \\ 
        \bottomrule[.1em]
    \end{tabular}
}
\caption{Performance of retrievers on \ours measured by nDCG@20 and MRR@20. } 
\label{tab: main-results-wo-instfol}
\end{table*}

\subsection{Detailed Results of Retrievers with Instructions as Prompts}
\label{sec: ab1}

As shown in ~\autoref{tab: Ab1 detail}, we present the instructions as prompts, as described in these papers. However, we use these instructions as prompts to differentiate from our instructions. The input to the retrievers should be formatted as ``[Prompt] [Query] [Instruction]." Additionally, there may be slight differences in the input format due to different models. 

\begin{table*}[htbp]
\centering
\scalebox{0.8}{
    \begin{tabular}{cccccccccc}
        \toprule[.1em]
        
          & FiQA & SciFact-open & NFCorpus & AILA & FIRE & TREC-PM & TREC-CDS & Average \\
        \midrule
        
        \multirow{2}{*}{Instructor-Base} & 0.392 & 0.451 & 0.482 & 0.059 & 0.506 & 0.174 & 0.091 & 0.308 \\ 
         & 0.393 & 0.445 & 0.489 & 0.059 & 0.499 & 0.232 & 0.080 & 0.314 \\ 
         \noalign{\vskip 0.5ex}\hdashline\noalign{\vskip 0.5ex}
         
        \multirow{2}{*}{Instructor-Large} & 0.488 & 0.463 & 0.564 & 0.070 & 0.510 & 0.149 & 0.167 & 0.345 \\ 
         & 0.493 & 0.469 & 0.567 & 0.072 & 0.516 & 0.166 & 0.155 & 0.348 \\ 
         \noalign{\vskip 0.5ex}\hdashline\noalign{\vskip 0.5ex}
         
        \multirow{2}{*}{Instructor-XL} & 0.484 & 0.488 & 0.530 & 0.071 & 0.529 & 0.169 & 0.188 & 0.351 \\ 
         & 0.489 & 0.489 & 0.544 & 0.072 & 0.533 & 0.204 & 0.205 & 0.362 \\ 
        \noalign{\vskip 0.5ex}\hdashline\noalign{\vskip 0.5ex}
        
        \multirow{2}{*}{E5-mistral-7b-inst} & 0.541 & 0.629 & 0.686 & 0.103 & 0.565 & 0.563 & 0.283 & 0.481 \\ 
        & 0.495 & 0.607 & 0.679 & 0.108 & 0.574 & 0.569 & 0.293 & 0.475 \\ 
        \noalign{\vskip 0.5ex}\hdashline\noalign{\vskip 0.5ex}
        
        \multirow{2}{*}{GritLM-7B} & 0.632 & 0.631 & 0.698 & 0.096 & 0.511 & 0.575 & 0.423 & 0.509 \\ 
         & 0.567 & 0.612 & 0.681 & 0.093 & 0.516 & 0.571 & 0.425 & 0.495 \\
        
        \bottomrule[.1em]
    \end{tabular}
}

\caption{Detailed nDCG@20 results of adding instructions as prompt. The first line is without instruction as a prompt, the second is with instruction as a prompt.}
\label{tab: Ab1 detail}
\end{table*}

\subsection{Detailed results of different retrievers on different levels. }
\label{sec: ab2}
The detailed result for each domain is shown in ~\autoref{tab: ab2 total}. The result of \ourmetric is shown in ~\autoref{tab: ab2-inst-total}.  From ~\autoref{tab: ab2-inst-total}, we can find that current information retrievers are not good at long instructions and instructions with highly dense expert knowledge.

\begin{table*}[htbp]
\centering
\scalebox{0.7}{
    \begin{tabular}{lcccccccccccc}
        \toprule[.1em]
          & \multicolumn{3}{c}{FiQA} & \multicolumn{3}{c}{AILA}  & \multicolumn{3}{c}{TREC-PM}  & \multicolumn{3}{c}{Scifact-open} \\ 
          \cmidrule(lr){2-4} \cmidrule(lr){5-7} \cmidrule(lr){8-10} \cmidrule(lr){11-13}
          & Level1 & Level2 & Level3 & Level1 & Level2 & Level3 & Level1 & Level2 & Level3 & Level1 & Level2 & Level3 \\ 
        \midrule
        BM25 & 0.282 & 0.221 & 0.239 & 0.158 & 0.060 & 0.030 & 0.505 & 0.437 & 0.482 & 0.568 & 0.438 & 0.480 \\
        Contriever & 0.146 & 0.121 & 0.111 & 0.144 & 0.018 & 0.012 & 0.112 & 0.084 & 0.077 & 0.306 & 0.252 & 0.325 \\
        ColBERT & 0.078 & 0.043 & 0.107 & 0.111 & 0.052 & 0.012 & 0.012 & 0.023 & 0.037 & 0.132 & 0.121 & 0.165 \\
        GTR-Base & 0.422 & 0.215 & 0.337 & 0.096 & 0.017 & 0.000 & 0.269 & 0.280 & 0.268 & 0.511 & 0.448 & 0.456 \\
        GTR-Large & 0.479 & 0.279 & 0.391 & 0.117 & 0.023 & 0.048 & 0.293 & 0.312 & 0.219 & 0.538 & 0.467 & 0.513 \\
        GTR-XL & 0.530 & 0.366 & 0.413 & 0.073 & 0.032 & 0.023 & 0.350 & 0.324 & 0.243 & 0.595 & 0.512 & 0.515 \\
        \midrule
        Instructor-Base & 0.424 & 0.361 & 0.387 & 0.110 & 0.024 & 0.000 & 0.119 & 0.208 & 0.197 & 0.481 & 0.429 & 0.449 \\
        Instructor-Large & 0.531 & 0.454 & 0.472 & 0.110 & 0.024 & 0.048 & 0.144 & 0.157 & 0.147 & 0.480 & 0.404 & 0.513 \\
        Instructor-XL & 0.558 & 0.435 & 0.445 & 0.122 & 0.012 & 0.042 & 0.190 & 0.181 & 0.135 & 0.536 & 0.461 & 0.476 \\
        E5-mistral-7b-inst & 0.628 & 0.490 & 0.488 & 0.162 & 0.041 & 0.060 & 0.582 & 0.567 & 0.537 & 0.673 & 0.620 & 0.601 \\
        GritLM-7B & 0.705 & 0.594 & 0.581 & 0.185 & 0.014 & 0.017 & 0.618 & 0.578 & 0.527 & 0.699 & 0.620 & 0.583 \\
        Promptriever-7B & 0.164 & 0.144 & 0.377 & 0.140 & 0.051 & 0.048 & 0.231 & 0.406 & 0.417 & 0.287 & 0.276 & 0.449 \\
        NV-Embed-v2 & 0.759 & 0.657 & 0.594 & 0.116 & 0.026 & 0.029 & 0.563 & 0.547 & 0.493 & 0.714 & 0.634 & 0.625 \\
        \midrule
        OpenAI-v3-small & 0.529 & 0.410 & 0.429 & 0.148 & 0.032 & 0.013 & 0.428 & 0.436 & 0.371 & 0.631 & 0.545 & 0.563 \\
        OpenAI-v3-large & 0.616 & 0.488 & 0.511 & 0.158 & 0.056 & 0.062 & 0.553 & 0.523 & 0.483 & 0.621 & 0.578 & 0.587 \\
        
        \bottomrule[.1em]
    \end{tabular}
}

\caption{Detailed nDCG@20 results of different retrievers on different levels. }
\label{tab: ab2 total}
\end{table*}

\begin{table*}[htbp]
\centering
\scalebox{0.7}{
    \begin{tabular}{lcccccccccccc}
        \toprule[.1em]
          & \multicolumn{3}{c}{FiQA} & \multicolumn{3}{c}{AILA}  & \multicolumn{3}{c}{TREC-PM}  & \multicolumn{3}{c}{Scifact-open} \\
          \cmidrule(lr){2-4} \cmidrule(lr){5-7} \cmidrule(lr){8-10} \cmidrule(lr){11-13}
          & Level1 & Level2 & Level3 & Level1 & Level2 & Level3 & Level1 & Level2 & Level3 & Level1 & Level2 & Level3 \\ 
        \midrule
        BM25 & -3.76 & -2.48 & 9.53 & -0.08 & -0.15 & 0.29 & 0.83 & 2.28 & 7.17 & -0.70 & -5.36 & 4.85 \\ 
        Contriever & -2.04 & -0.69 & 4.28 & 0.07 & -0.11 & 0.30 & 0.07 & 1.91 & 1.75 & -3.20 & -11.25 & -10.21 \\ 
        ColBERT & -4.43 & -7.28 & 12.21 & 0.24 & 0.85 & -1.13 & 0.14 & 1.11 & 1.84 & -0.23 & -1.06 & 2.31 \\ 
        GTR-Base & -3.76 & -12.39 & 2.30 & -0.06 & 0.51 & 1.18 & -4.56 & -0.12 & 2.73 & -1.09 & -0.39 & -0.39 \\ 
        GTR-Large & -6.40 & -13.20 & 0.18 & 0.12 & -0.31 & -0.88 & -3.37 & -0.28 & -0.73 & -0.18 & 0.64 & -0.30 \\ 
        GTR-XL & -4.45 & -10.59 & 0.49 & 0.15 & -0.20 & -0.34 & -3.21 & -0.73 & 1.40 & -0.14 & 0.46 & -9.12 \\ 
        \midrule
        Instructor-Base & -1.43 & 1.52 & 9.83 & 0.16 & -0.23 & 0.61 & -4.31 & 1.64 & 6.70 & -0.94 & 0.66 & 1.54 \\ 
        Instructor-Large & -0.34 & 2.49 & 8.79 & -0.22 & 0.04 & 1.04 & -2.38 & -2.77 & -6.42 & -1.15 & -2.07 & 3.83 \\ 
        Instructor-XL & -0.66 & -1.93 & 5.67 & 0.03 & -0.07 & -0.86 & -2.00 & -0.82 & -3.37 & 0.53 & 2.34 & -9.95 \\ 
        E5-mistral-7b-inst & 0.14 & -0.13 & 12.78 & 1.11 & -0.08 & -0.79 & -0.89 & 1.10 & 2.55 & -0.22 & 0.19 & 0.18 \\ 
        GritLM-7B & 0.25 & 1.68 & 7.32 & -0.37 & 0.06 & -0.66 & -0.75 & 0.14 & 0.33 & -0.36 & 0.92 & -0.73 \\ 
        Promptriever-7B & -2.41 & 2.18 & 27.09 & -1.22 & 0.37 & -0.07 & 2.09 & 12.49 & 25.21 & -2.02 & -5.17 & 18.27 \\ 
        NV-Embed-v2 & 0.14 & 1.13 & 7.02 & -0.83 & -0.20 & -0.02 & 0.37 & 0.44 & 1.35 & -0.54 & 0.51 & -3.27 \\ 
        \midrule
        OpenAI-v3-small & -0.60 & -0.62 & 8.16 & 0.12 & 0.10 & -1.09 & -1.93 & -0.80 & -0.88 & 0.26 & 1.72 & -4.79 \\ 
        OpenAI-v3-large & -0.95 & -0.97 & 6.63 & 0.54 & -0.12 & -0.50 & -1.72 & -1.37 & 3.64 & -0.19 & 0.75 & -1.99 \\ 
        
        \bottomrule[.1em]
    \end{tabular}
}

\caption{Detailed \ourmetric@20 results of different retrievers on different levels. }
\label{tab: ab2-inst-total}
\end{table*}

\end{document}